\documentclass[authoryear]{eshbaugh}
\usepackage{multirow}
\usepackage{float}

\title{``Where Can I Trust You?'': Boundary-Aware Evaluation of Surrogate Fidelity}

\author{Jackson Eshbaugh}
\affiliation{Department of Computer Science\\Lafayette College\\Easton, PA, USA}
\email{eshbaugj@lafayette.edu}
\orcid{0009-0009-1806-2166}

\publicationstatus{Preprint}
\version{1.0}
\preprintdate{September 2026}

\providecommand{\citet}{\textcite}
\providecommand{\citep}{\parencite}

\begin{document}

\maketitle

\begin{abstract}
Surrogate models are commonly evaluated by how often they agree with their teacher model over an evaluation set. 
Local variation in this agreement is well known, but its structure and consequences are less clear.
We ask whether disagreement is systematically concentrated near the teacher's decision boundary and whether retaining that structure provides information beyond a single global score.
Across several datasets and surrogate model classes, we find substantially lower fidelity near teacher decision boundaries under two different methods of identifying near-boundary examples.
Moreover, conditioning agreement on confidence-defined regions improves prediction of teacher--surrogate agreement when evaluation-set composition changes, relative to the global score alone.
Yet surrogates that agree equally well with the teacher both globally and near the decision boundary can respond very differently to changes selected using the surrogate itself.
Finally, we show that independently trained deep teachers can agree on most predictions while identifying different examples as lying near their decision boundaries, complicating the use of those boundaries as stable reference regions for evaluating surrogates.
Together, these results show that surrogate fidelity depends not only on how often a surrogate agrees with its teacher, but also on where that agreement holds and, for deep models, how stable the teacher's decision boundary is across training runs.
\end{abstract}

\section{Introduction}
\label{sec:intro}

Surrogate models are only as useful as their ability to reliably stand in for the models they approximate.
This reliability is typically evaluated through fidelity: the agreement between a surrogate and its teacher over an evaluation set.
Prior work has shown that this agreement can vary substantially across the evaluation space.
What remains less clear is whether that variation has systematic structure and whether retaining that structure provides information beyond a single global score.

In this work, we study the structure of teacher--surrogate disagreement for a fixed pair, with particular attention to the teacher's decision boundary.
We ask whether disagreement is systematically concentrated near the teacher's decision boundary,
whether similar levels of global and boundary-local fidelity correspond to similar counterfactual transfer rates, and
whether measuring agreement across teacher-confidence levels better predicts fidelity under changes in the evaluation composition.
We further examine whether the boundary regions used for this analysis are stable across independently trained deep teachers.

We characterize fidelity as a conditional property of a teacher--surrogate pair over an evaluation space, rather than as a single global quantity.
Our contributions are:
(1) characterizing how teacher--surrogate disagreement varies with proximity to the teacher's decision boundary, using confidence-based and geometric proxies;
(2) evaluating whether confidence-stratified profiles predict fidelity better than a global score when the evaluation composition changes;
(3) examining what matched global and boundary-local fidelity does and does not establish about example-based counterfactual transfer; and
(4) studying whether independently trained deep teachers identify the same examples as lying near their decision boundaries.

\section{Related Work}
\label{sec:RW}

Fidelity arises from a broader problem of determining how well one model preserves the behavior of another.
In knowledge distillation, compact student models are trained to reproduce the outputs or intermediate representations of larger teacher models \citep{hintonDistillingKnowledge2015,romeroFitNetsHints2015}.
Surrogate-based interpretability uses a teacher--student setup for a different purpose:
a simpler, more interpretable model is trained to approximate the behavior of an otherwise difficult-to-interpret predictor \citep{ribeiroWhyShouldTrust2016,lyuFaithfulModelExplanation2024}.

The quality of this approximation is commonly evaluated through fidelity, reported as an aggregate measure of agreement between a surrogate and its teacher over an evaluation set \citep{setzuGLocalX2021,confalonieriUsingOntologies2021, henckaertsWhenStakes2022,mariottiBeyondPredictionSimilarity2023}.
Such aggregation can obscure heterogeneity in approximation quality. \citet{balagopalanRoadExplainability2022}, for example, show that both local and global explanation models can exhibit substantially different fidelity across protected subgroups, so that average fidelity may overstate explanation quality for some populations.
Their analysis conditions fidelity on externally defined group membership.
In contrast, we define regions using the teacher's own decision structure and ask whether disagreement is systematically concentrated near its decision boundary.

Locality also plays a central role in how surrogate explanations are constructed and evaluated.
\citet{ribeiroWhyShouldTrust2016} fit interpretable surrogates in neighborhoods around individual predictions.
\citet{laugelDefiningLocality2018} show that the choice of neighborhood is important for accurately approximating local decision-boundary structure, while \citet{poyiadziUnderstandingSurrogateExplanations2021} subsequently characterize locality as a continuum of coverage, studying how the fidelity--complexity tradeoff changes as the region used to fit the surrogate contracts.
More recently, \citet{huynhGloballyValid2026} distinguish local from global validation contexts, showing that the measured fidelity of an individual rule explanation can decrease when the same rule is evaluated beyond its original neighborhood.

Related work has also questioned what high predictive fidelity actually establishes about a surrogate.
\citet{mariottiBeyondPredictionSimilarity2023} show that models with high prediction-level fidelity can nevertheless differ substantially in feature-attribution structure, motivating ShapGAP as a comparison of model explanations rather than outputs alone.
Moreover, \citet{eshbaughFaithfulWhat2026} distinguishes fidelity to a learned model from recovery of the task-relevant structure underlying predictive performance, showing that a surrogate can closely approximate a neural network's learned function while failing to preserve its predictive advantage.
The present work addresses a complementary limitation: for a fixed teacher--surrogate pair, we ask whether predictive disagreement is systematically organized relative to the teacher's decision boundary and whether retaining that structure improves prediction of fidelity when evaluation composition changes.
We further examine whether matching both global and boundary-local agreement is sufficient to produce similar counterfactual transfer behavior.

Prior work has examined the reproducibility of neural-network decision boundaries across independent training runs.
\citet{somepalliCanNeural2022} find substantial similarity in decision regions across random initializations of the same architecture, particularly for wide convolutional networks, while also observing systematic differences across model families.
\citet{leiUnderstandingDeep2025} study decision-boundary variability across training repeats, operationalizing it through predictive disagreement between independently trained networks and relating lower variability to improved generalization.
Our analysis concerns a different notion of boundary stability: whether independently trained teachers identify the same evaluation examples as lying near their decision boundaries.

These studies motivate our focus on how teacher--surrogate disagreement is distributed relative to the teacher's decision boundary,
whether that structure improves prediction of fidelity when evaluation composition changes, 
what matched global and boundary-local agreement establishes about counterfactual transfer, and 
how stable the relevant boundary regions are across independently trained teachers.

\section{Methodology}
\label{sec:methodology}

Our methodology is designed to characterize teacher--surrogate agreement beyond a single global fidelity score.
We first define global and boundary-conditioned fidelity under two notions of boundary proximity, then describe protocols for predicting fidelity under changes in evaluation composition and for measuring example-based counterfactual transfer.
Finally, we describe our analysis of boundary stability across independently trained deep teachers.

\subsection{Teacher--Surrogate Fidelity}

Let \( T \) represent a teacher model and \( S_T \) a surrogate model trained on input--output pairs \( (x, T(x)) \) to reproduce the teacher's predicted class rather than the ground-truth label.
Fidelity is measured over an evaluation set \( X = \{x_i\}_{i=1}^n \) and quantifies agreement with the teacher rather than accuracy against ground-truth labels.
Global fidelity is defined in Equation~\ref{eq:global_fid}.
This scalar gives the fraction of evaluation examples on which the surrogate and teacher make the same prediction.

\begin{equation}
    \label{eq:global_fid}
    F(T,S_T;X)
    =
    \frac{1}{|X|}
    \sum_{x\in X}
    \mathbf{1}[T(x)=S_T(x)]
\end{equation}

To examine this agreement, our tabular experiments use a common teacher--surrogate training setup.
Features are standardized using statistics estimated from the training partition only.
Teachers are binary multilayer perceptrons with hidden widths 64 and 32 and ReLU activations, trained on ground-truth labels using binary cross-entropy.
Our primary capacity analyses use decision-tree surrogates; in Section~\ref{sec:boundary_results}, we additionally evaluate logistic-regression, \( k \)-nearest-neighbor, and multilayer-perceptron surrogates to test whether boundary-local disagreement is specific to the decision-tree model class.
Experiments are repeated across ten random seeds unless otherwise stated.
Full dataset, preprocessing, model, training, and boundary-computation details are provided in Appendix~\ref{app:experimental_details}.

\subsection{Boundary-Aware Fidelity}

Global fidelity does not distinguish between disagreement occurring in different regions of the evaluation space.
We therefore condition fidelity on proximity to the teacher's decision boundary.

\paragraph{Confidence-based proximity.}
For a binary teacher, let \( f_T(x)\in \mathbb{R} \) denote the teacher logit, such that
\(
T(x)=\mathbf{1}[f_T(x)\geq 0].
\)

Our first measure uses the magnitude of the teacher logit, \( c_T(x)=|f_T(x)|. \)
Smaller values indicate lower teacher confidence and serve as a proxy for greater proximity to the teacher's prediction transition.
For a boundary fraction \( \alpha\in(0,1] \), we define \( B_{\alpha}^{\mathrm{conf}}(T;X) \) as the \( \lceil \alpha|X| \rceil \) evaluation examples with the smallest values of \( c_T(x) \).

\paragraph{Segment-crossing proximity.} 
Because confidence-based proximity depends directly on the teacher's output score, we also use a geometric proxy that depends only on teacher predictions and the geometry of the evaluation set.
For each \( x\in X \), we identify the nearest observed evaluation example \( x'\in X \) for which \( T(x')\neq T(x) \), using Euclidean distance in the standardized feature space.
We then perform a binary search along the line segment connecting \( x \) and \( x' \) to approximate a point at which the teacher prediction changes.
The distance from \( x \) to this estimated crossing defines \( d_T^{\mathrm{seg}}(x) \). This quantity is an approximate segment-crossing distance and should not be interpreted as the exact minimum Euclidean distance to the teacher's decision boundary.
We define \( B_{\alpha}^{\mathrm{seg}}(T;X) \) as the \( \lceil\alpha|X|\rceil \) examples with the smallest values of \( d_T^{\mathrm{seg}}(x) \).

For either boundary definition \(B_{\alpha}\), boundary-conditioned fidelity is given in Equation~\ref{eq:boundary_fid}. Equation~\ref{eq:boundary_gap} reports the corresponding boundary gap where positive values indicate lower fidelity on the selected examples than on the evaluation set as a whole.

\begin{equation}
    \label{eq:boundary_fid}
    F_{\alpha}(T,S_T;X)
    =
    \frac{1}{|B_{\alpha}|}
    \sum_{x\in B_{\alpha}}
    \mathbf{1}[T(x)=S_T(x)]
\end{equation}

\begin{equation}
    \label{eq:boundary_gap}
    \Delta_{\alpha}
    =
    F(T,S_T;X)
    -
    F_{\alpha}(T,S_T;X)
\end{equation}

We use these measures to examine how disagreement near the boundary varies with region size, surrogate capacity, and surrogate model class.
For the boundary-local analysis, we evaluate boundary fractions \( \alpha\in\{0.10,0.20,0.25,0.40\} \).
Our primary capacity analysis uses decision-tree surrogates with maximum depths \( \{1,2,3,5,7,10\} \).
To test whether the observed boundary-local structure is specific to decision trees, we additionally compare representative decision-tree, logistic-regression, \( k \)-nearest-neighbor, and multilayer-perceptron surrogates at \( \alpha=0.25 \).
These configurations are a depth-5 decision tree, logistic regression with \( C=1 \), \( k \)-NN with \( k = 5 \), and a one-hidden-layer MLP with width 32.

\subsection{Predicting Fidelity Under Changes in Evaluation Composition}

Having defined fidelity within selected regions, we examine whether measuring agreement across teacher-confidence levels improves predictions of fidelity when evaluation composition changes.
For each tabular dataset, we reserve \( 60\% \) of the data for training and \( 40\% \) for evaluation.
The evaluation partition is then divided evenly into disjoint source and target pools, stratified by the teacher's prediction.
The source pool is used to estimate agreement within confidence bins; the target pool provides the bin weights used to predict target fidelity and the observed agreement used to evaluate those predictions.

We partition the source pool into three quantile bins according to teacher confidence
\( c_T(x)=|f_T(x)| \). For each bin \( b \), let \( X_b \) denote the source-pool examples assigned to that bin.
We estimate conditional fidelity as
\( F_b=\frac{1}{|X_b|}\sum_{x\in X_b}\mathbf{1}[T(x)=S_T(x)] \). The source-derived bin boundaries are then applied unchanged to the target examples.
Let \( w_b=\frac{|X_{\mathrm{target},b}|}{|X_{\mathrm{target}}|} \) denote the proportion of target examples assigned to bin \( b \).

We predict target fidelity as
\begin{equation}
    \hat{F}_{\mathrm{profile}}
    =
    \sum_b w_b F_b.
\end{equation}

This prediction weights source-bin agreement rates by the composition of the target pool.
We compare it with a baseline using source-pool global fidelity directly, \( \hat{F}_{\mathrm{global}}=F(T,S_T;X_{\mathrm{source}}) \).

To compare these predictors under changes in evaluation composition, we divide the target pool at its median teacher confidence into lower- and higher-confidence subsets.
We treat each subset as a separate target distribution and apply both fidelity predictors to each.
Prediction error is measured as the absolute difference between predicted and observed target fidelity.
Our primary analysis uses three confidence bins; we additionally evaluate five bins as a sensitivity analysis.

\subsection{Example-Based Counterfactual Transfer}
Beyond predicting fidelity under changes in evaluation composition, we examine whether teacher--surrogate agreement corresponds to similar behavior under changes selected using the surrogate.
We use an example-based counterfactual transfer protocol with disjoint training, candidate, and test partitions containing \( 60\% \), \( 20\% \), and \( 20\% \) of the data, respectively.

For a test example \( x \), we consider only cases in which the teacher and surrogate initially agree, \( T(x)=S_T(x) \).
Let \( y' \) denote the class opposite to the surrogate's prediction on \( x \).
From the candidate set \( X_\text{candidate} \), we select the nearest example
\begin{equation}
    x^\star
    =
    \operatorname*{arg\,min}_{z\in X_{\text{candidate}}}
    \|x-z\|_2
    \quad
    \text{subject to}
    \quad
    S_T(z)=y',
\end{equation}
where distance is measured in the standardized feature space.
The selected candidate therefore represents the nearest available example that changes the surrogate's prediction. 
Counterfactual transfer occurs when the teacher also assigns that target class to the selected candidate, \( T(x^\star)=y' \).

We define the transfer rate as the fraction of eligible test examples for which this condition holds.
This protocol measures whether a surrogate-selected change that alters the surrogate's prediction also induces the corresponding change in the teacher; it does not assume that the selected example is a causal or actionable counterfactual.

To examine whether observational fidelity is sufficient to characterize transfer behavior, we compare surrogate configurations with similar global fidelity and boundary-conditioned fidelity at \( \alpha = 0.25 \).

\subsection{Boundary Stability Across Deep Teachers}
Boundary-aware fidelity is defined relative to a teacher, but in deep learning the teacher itself may vary across independent training runs.
We therefore measure whether independently trained teachers identify the same examples as lying near their decision boundaries.

For a multiclass teacher \( T \), let \( f_T(x)\in\mathbb{R}^K \) denote its logit vector.
Let \(f_T^{(1)}(x)\) and \(f_T^{(2)}(x)\) denote its largest and second-largest logits, respectively.
We define the teacher's prediction margin as
\begin{equation}
    m_T(x)
    =
    f_T^{(1)}(x)-f_T^{(2)}(x).
\end{equation}
Smaller margins indicate greater proximity to a prediction transition.

Because absolute logit scales are not necessarily comparable across independently trained networks, we define boundary regions by rank rather than by a common margin threshold.
For a boundary fraction \(\alpha\), the set \(B_\alpha(T;X)\) contains the \(\lceil \alpha|X| \rceil\) examples with the smallest top-two logit margins for teacher \(T\).

We train five independently trained teachers on MNIST~\citep{lecun1998gradient} and CIFAR-10~\citep{krizhevsky2009learning} and compare every unique teacher pair.

For teachers \( T_i \) and \( T_j \), boundary-set stability is measured using Jaccard overlap (Equation~\ref{eq:jaccard}).
\begin{equation}
\label{eq:jaccard}
    J_\alpha(T_i,T_j)
    =
    \frac{
        |B_\alpha(T_i;X)\cap B_\alpha(T_j;X)|
    }{
        |B_\alpha(T_i;X)\cup B_\alpha(T_j;X)|
    }.
\end{equation}

We additionally measure pairwise predictive agreement globally and over the union and intersection of the two teachers' boundary regions.
These comparisons are reported descriptively across teacher pairs, because individual teachers participate in multiple pairs so the pairwise observations are not independent.

\section{Results}
\label{sec:results}

Using the methods described in Section~\ref{sec:methodology}, we examine where teacher-surrogate disagreement occurs, whether confidence-stratified fidelity improves prediction when evaluation composition changes, and what global and boundary-local agreement establish about counterfactual transfer.
We then assess how stable the boundary regions are across independently trained teachers.

\subsection{Boundary-Local Structure of Surrogate Disagreement}
\label{sec:boundary_results}

To examine whether disagreement is concentrated near the teacher's decision boundary, we compare global and boundary-local fidelity across a range of decision-tree surrogate capacities.
Boundary-local fidelity is consistently lower than global fidelity on three datasets---Breast Cancer~\citep{wolberg1993wdbc}, Diabetes~\citep{efron2004least}, and Wine~\citep{aeberhard1992wine}---while Iris~\citep{fisher1936iris} exhibits little difference between the two.
At a boundary fraction of \(25\%\), global fidelity exceeds confidence-based boundary fidelity by about 12, 19, and 10 percentage points (pp) on Breast Cancer, Diabetes, and Wine, respectively, compared with 1.3 pp on Iris.

\begin{figure}[t]
    \centering
    \includegraphics[width=0.8\linewidth]{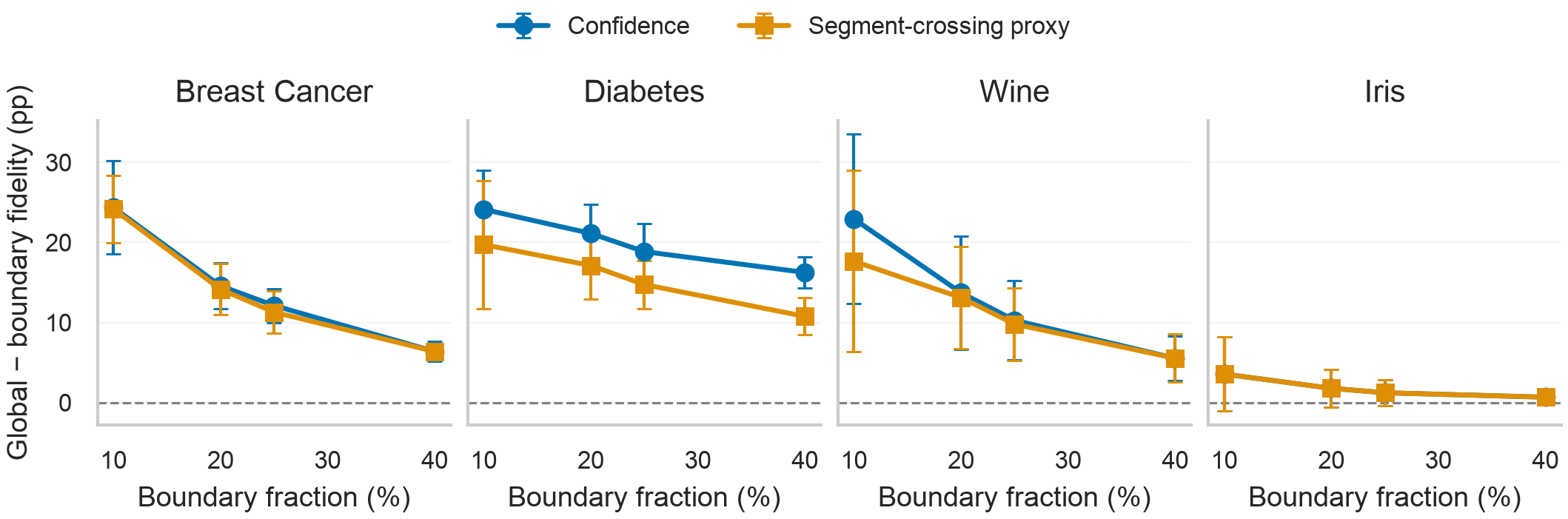}
    \caption{Global-to-boundary fidelity gaps, in percentage points, as the boundary fraction varies under confidence-based and approximate segment-crossing measures.
    Points show mean gaps across random seeds after averaging over decision-tree depths within each seed; error bars show approximate 95\% confidence intervals across seeds.}
    \label{fig:boundary_fidelity}
\end{figure}

Figure~\ref{fig:boundary_fidelity} shows that this gap widens as evaluation is restricted to examples closer to the teacher's decision boundary.
The same qualitative pattern appears under both the confidence-based and segment-crossing measures, suggesting that the observed concentration of disagreement is not specific to a single way of identifying examples near the boundary.

We next test whether this pattern is specific to decision-tree surrogates. Table~\ref{tab:surrogate_family_robustness} compares representative decision-tree, logistic-regression, \( k \)-nearest-neighbor, and MLP surrogates at a fixed boundary fraction of \( 25\% \).
On Breast Cancer, mean boundary-local fidelity is lower than global fidelity for all four surrogate families under both measures, with smaller gaps for the higher-fidelity logistic-regression and MLP surrogates.
The result is particularly pronounced on Diabetes: confidence-based boundary fidelity is lower than global fidelity by approximately \( 19 \)--\( 22 \) pp across all four surrogate families, with a positive gap in every random seed.

\begin{table*}[t]
\centering

\begin{tabular}{llccccc}
\toprule
& &
\multirow{2}{*}{\shortstack{Global\\fidelity (\%)}} &
\multicolumn{2}{c}{Confidence-based boundary} &
\multicolumn{2}{c}{Segment-crossing boundary} \\
\cmidrule(lr){4-5}
\cmidrule(lr){6-7}
Dataset &
Surrogate &
&
\shortstack{Gap (pp)\\mean $\pm$ 95\% CI} &
Pos. &
\shortstack{Gap (pp)\\mean $\pm$ 95\% CI} &
Pos. \\
\midrule

\multirow{4}{*}{Breast Cancer}
& Decision tree & 94.0 & 11.9 $\pm$ 2.2 & 10/10 & 10.1 $\pm$ 2.8 & 10/10 \\
& Logistic regression & 98.2 & 5.2 $\pm$ 1.6 & 10/10 & 5.2 $\pm$ 1.6 & 10/10 \\
& \(k\)-NN & 96.4 & 10.6 $\pm$ 3.4 & 10/10 & 9.9 $\pm$ 3.6 & 10/10 \\
& MLP & 98.8 & 3.4 $\pm$ 1.7 & 8/10 & 3.2 $\pm$ 1.6 & 8/10 \\

\midrule

\multirow{4}{*}{Diabetes}
& Decision tree & 73.6 & 18.6 $\pm$ 4.2 & 10/10 & 13.9 $\pm$ 4.6 & 9/10 \\
& Logistic regression & 79.5 & 20.7 $\pm$ 3.4 & 10/10 & 16.2 $\pm$ 3.7 & 10/10 \\
& \(k\)-NN & 80.7 & 20.4 $\pm$ 4.7 & 10/10 & 15.7 $\pm$ 4.3 & 10/10 \\
& MLP & 82.9 & 22.3 $\pm$ 4.1 & 10/10 & 13.8 $\pm$ 3.4 & 10/10 \\

\midrule

\multirow{4}{*}{Wine}
& Decision tree & 95.0 & 11.4 $\pm$ 5.2 & 9/10 & 11.4 $\pm$ 4.4 & 9/10 \\
& Logistic regression & 99.4 & 1.6 $\pm$ 2.2 & 2/10 & 1.6 $\pm$ 2.2 & 2/10 \\
& \(k\)-NN & 98.5 & 3.5 $\pm$ 3.1 & 5/10 & 3.5 $\pm$ 3.1 & 5/10 \\
& MLP & 99.6 & 1.1 $\pm$ 1.4 & 2/10 & 0.3 $\pm$ 1.1 & 1/10 \\

\midrule

\multirow{4}{*}{Iris}
& Decision tree & 99.6 & 1.2 $\pm$ 1.6 & 2/10 & 1.2 $\pm$ 1.6 & 2/10 \\
& Logistic regression & 99.6 & 1.2 $\pm$ 1.6 & 2/10 & 1.2 $\pm$ 1.6 & 2/10 \\
& \(k\)-NN & 99.3 & 1.8 $\pm$ 1.8 & 3/10 & 1.8 $\pm$ 1.8 & 3/10 \\
& MLP & 99.6 & 1.2 $\pm$ 1.6 & 2/10 & 1.2 $\pm$ 1.6 & 2/10 \\

\bottomrule
\end{tabular}
\caption{Boundary-local fidelity degradation across surrogate families at \(\alpha = 0.25\).
Global fidelity and boundary gaps are averaged over ten seeds; gaps are reported in percentage points with approximate 95\% confidence intervals.
``Pos.'' counts seeds for which global fidelity exceeds boundary-local fidelity.}
\label{tab:surrogate_family_robustness}
\end{table*}

Wine provides an informative contrast. The decision-tree surrogate exhibits a substantial boundary-local gap, while logistic regression, \( k \)-NN, and the MLP achieve approximately \( 98.5\% \)--\( 99.6\% \) global fidelity and show much smaller degradation.
Thus, the boundary-local concentration of disagreement is not an artifact of decision-tree surrogates, although its magnitude is smaller when the surrogate reproduces the teacher almost everywhere.
More broadly, globally aggregated fidelity can conceal systematic spatial structure in teacher--surrogate disagreement across substantially different surrogate model classes.
Additional decision-tree capacity results and surrogate-family robustness analyses are reported in Appendix~\ref{app:boundary_results}.

\subsection{Predicting Fidelity Under Changes in Evaluation Composition}
\label{sec:shift_results}

The concentration of disagreement near the teacher's decision boundary suggests that conditioning on teacher confidence may improve prediction of surrogate fidelity when evaluation composition changes.
We compare a global-fidelity baseline with a confidence-stratified profile that predicts fidelity on the target set based on its composition across confidence regions.
Across 180 evaluation conditions with altered composition, the global baseline yields a mean absolute error of \( 7.67 \) pp, compared with \( 4.62 \) pp for the three-bin stratified profile, a \( 39.7\% \) reduction in error.

\begin{figure}[t]
    \centering
    \includegraphics[width=0.8\linewidth]{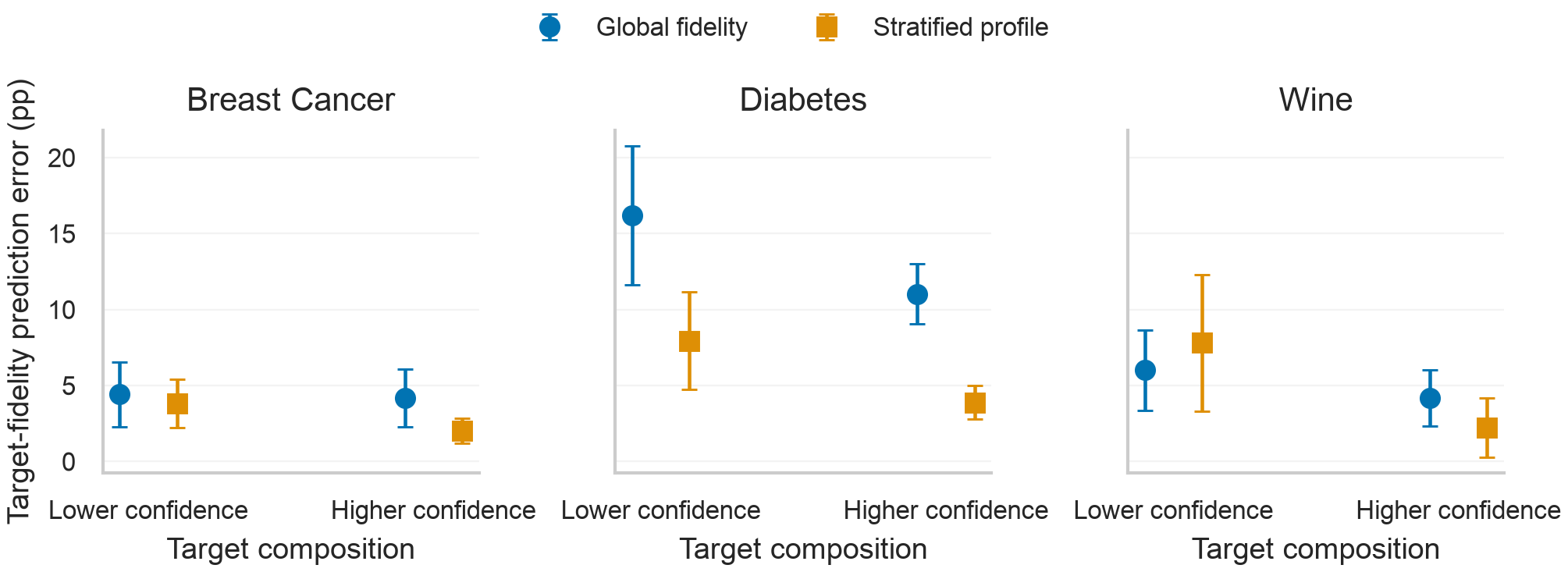}
    \caption{Confidence-stratified fidelity improves prediction when evaluation composition changes.
    Mean absolute error in predicting target-set fidelity under shifts toward lower- and higher-confidence regions.
    Predictions using a three-bin confidence-stratified fidelity profile are compared with a single global-fidelity baseline.
    Error bars indicate 95\% confidence intervals across random seeds, after averaging over surrogate depths within each seed.}
    \label{fig:distribution_shift}
\end{figure}

Figure~\ref{fig:distribution_shift} shows that the improvement is especially pronounced under shifts toward higher-confidence regions, while gains under shifts toward lower-confidence regions are more modest.
The stratified profile outperforms the global baseline in \( 121 \) of \( 180 \) conditions, with \( 9 \) ties.
At the dataset level, mean absolute error decreases from \( 4.30 \) pp to \( 2.93 \) pp on Breast Cancer and from \( 13.62 \) pp to \( 5.92 \) pp on Diabetes, while remaining nearly unchanged on Wine (\( 5.09 \) pp versus \( 5.02 \) pp).
A complementary analysis across representative decision-tree, logistic-regression, \( k \)-NN, and MLP surrogates yields the same overall pattern: mean absolute prediction error decreases from \( 6.79 \) pp to \( 3.72 \) pp, with lower mean error for the stratified profile in each surrogate family.
These results show that agreement measured across teacher-confidence regions provides predictive information about target fidelity beyond the source-set global score when evaluation composition changes.
Additional sensitivity analyses for the number of confidence bins and surrogate-family robustness are reported in Appendix~\ref{app:shift_results}.

\subsection{Counterfactual Transfer Under Matched Fidelity}
\label{sec:counterfactual_results}

The predictive value of boundary-aware fidelity does not mean that observational agreement is sufficient to characterize downstream behavior.
We therefore evaluate example-based counterfactual transfer: whether a counterfactual selected using the surrogate induces the same target-class change in the teacher.
On Breast Cancer, two surrogates can have identical global fidelity (\( 94.74\% \)) and identical boundary-conditioned fidelity (\( 82.76\% \)) while exhibiting sharply different counterfactual transfer rates: \( 73.15\% \) versus \( 24.07\% \).
Thus, matching both global and boundary-local observational agreement does not guarantee matched behavior under surrogate-selected changes.

\begin{figure}[t]
    \centering
    \includegraphics[width=0.6\linewidth]{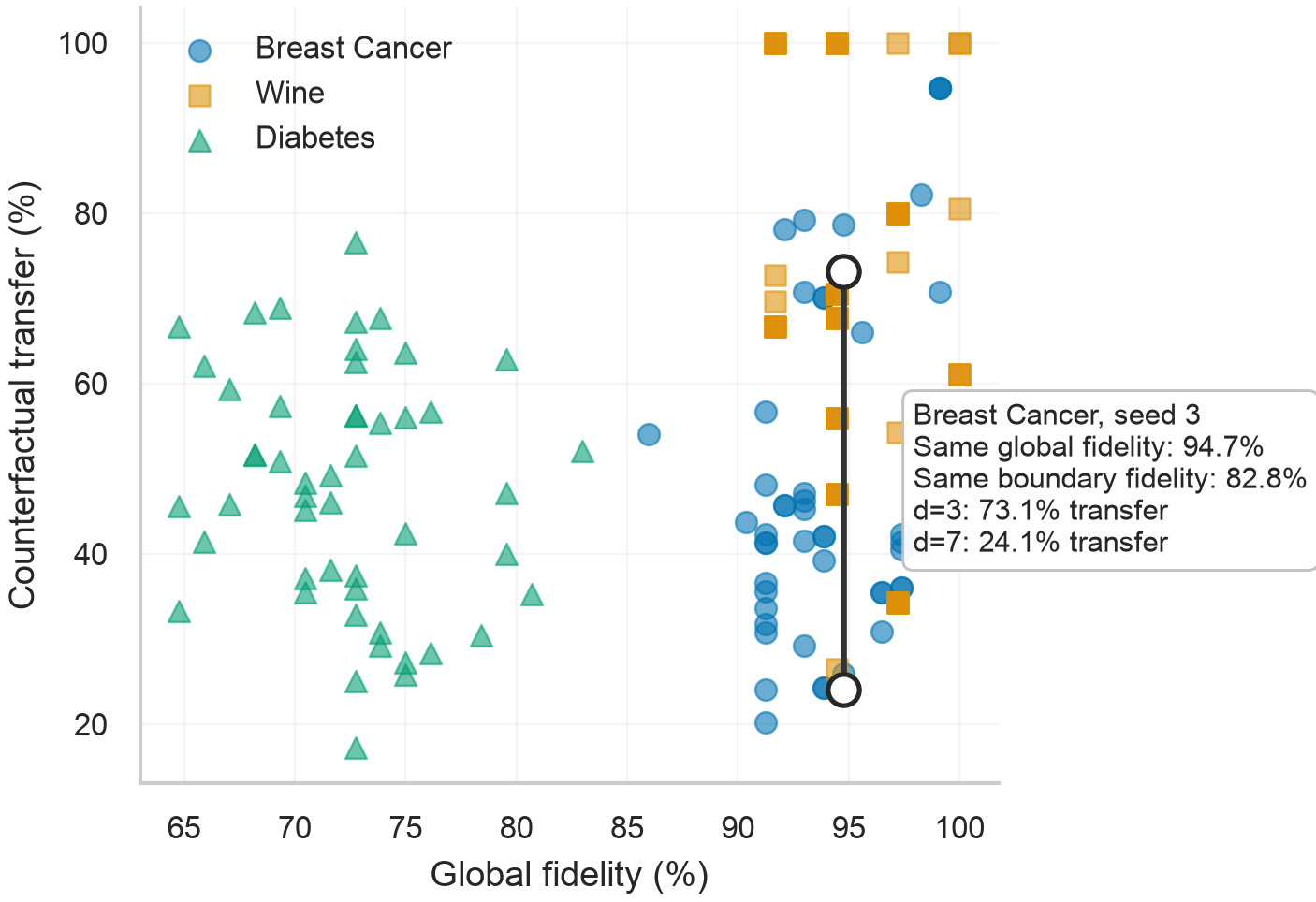}
    \caption{Similar observational agreement can correspond to substantial differences in counterfactual transfer.
    Each point represents one surrogate configuration, with global fidelity on the \( x \)-axis and counterfactual transfer on the \( y \)-axis.
    The vertical spread at similar global fidelity indicates that aggregate agreement does not uniquely determine counterfactual behavior.
    The highlighted Breast Cancer pair has identical global and boundary-local fidelity, but counterfactual transfer rates of 73.1\% and 24.1\%.}
    \label{fig:counterfactual_transfer}
\end{figure}

Across datasets, surrogate configurations with similar global fidelity can exhibit substantial variation in counterfactual transfer, producing vertical spread in Figure~\ref{fig:counterfactual_transfer}.
The highlighted Breast Cancer pair provides the clearest controlled example because the two surrogates are matched on both global and boundary-conditioned fidelity.
These results show that observational agreement, even when localized, does not fully characterize how a surrogate will behave under changes selected using that surrogate.
Additional results across surrogate capacities and matched-fidelity surrogate pairs are reported in Appendix~\ref{app:counterfactual_results}.

\subsection{Boundary Stability Across Deep Teachers}
\label{sec:deep_results}

Boundary-aware evaluation treats the teacher's decision boundary as a meaningful reference structure.
In deep networks, however, independently trained teachers may make nearly identical predictions while identifying different examples as being near the decision boundary.
We observe this behavior on both MNIST~\citep{lecun1998gradient} and CIFAR-10~\citep{krizhevsky2009learning}.
At a boundary fraction of \(10\%\), independently trained MNIST teachers achieve approximately \(99.2\%\) pairwise predictive agreement, yet the sets of examples they identify as near the decision boundary have a mean Jaccard overlap of only \(39.4\%\).
For CIFAR-10, pairwise agreement remains approximately \(96.0\%\), while mean boundary-set overlap is only \(44.3\%\).

\begin{figure}[t]
    \centering
    \includegraphics[width=0.8\linewidth]{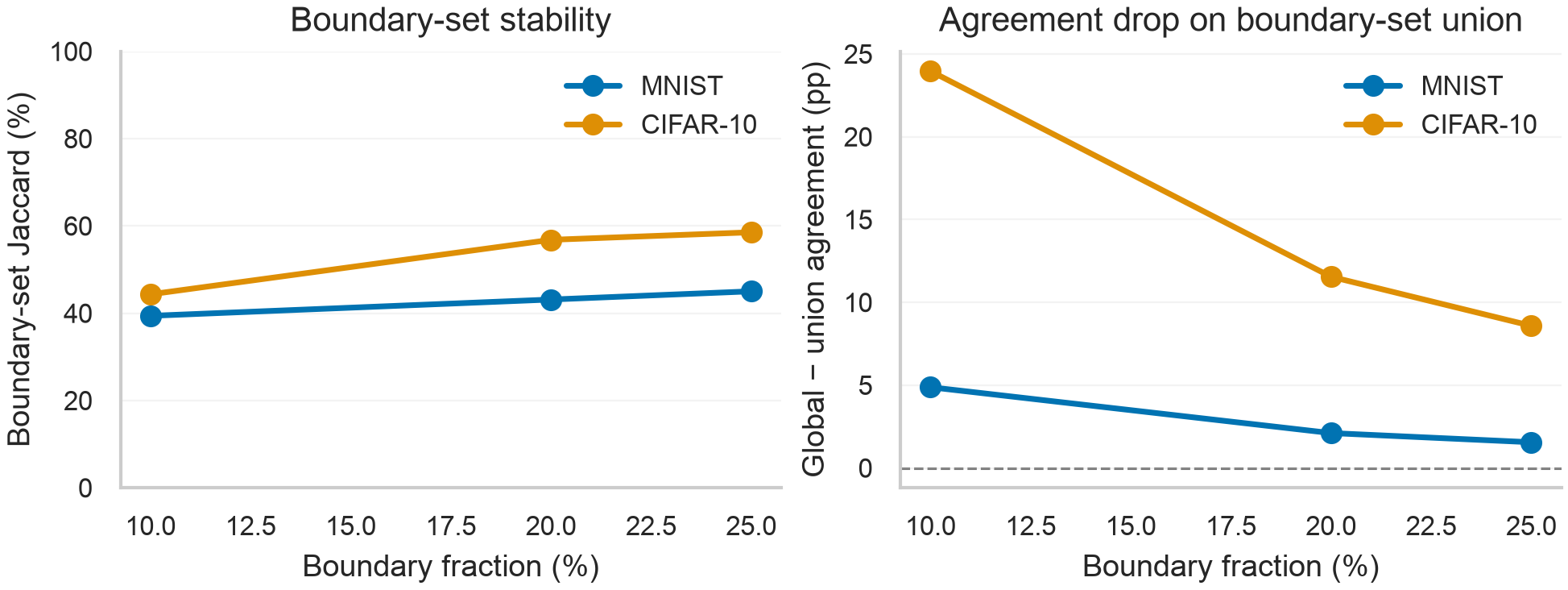}
    \caption{
Boundary regions are only partially stable across independently trained deep teachers.
Left: mean Jaccard overlap between teachers' lowest-margin examples as the boundary fraction varies.
Right: the drop from global pairwise predictive agreement to agreement over the union of the two teachers' boundary regions.
Results are shown for independently trained teachers on MNIST and CIFAR-10.
}
    \label{fig:deep_boundary_stability}
\end{figure}

Figure~\ref{fig:deep_boundary_stability} further shows that this partial boundary-set overlap is accompanied by lower predictive agreement within those regions.
On CIFAR-10, predictive agreement drops from approximately \( 96.0\% \) globally to \( 72.1\% \) over the union of the two teachers' tightest boundary regions, a decrease of roughly \( 24 \) percentage points.
The corresponding drop on MNIST is smaller but remains concentrated near the boundary.
These results suggest that boundary-aware fidelity in deep networks should be interpreted relative to the particular teacher being evaluated unless the relevant boundary regions are stable across independent training runs.
Individual teacher accuracies and additional pairwise boundary-set stability statistics are reported in Appendix~\ref{app:deep_results}.

\section{Discussion}
\label{sec:disc}

Teacher--surrogate disagreement is systematically organized relative to the teacher's decision boundary, and retaining this structure improves prediction of surrogate fidelity when evaluation composition changes.
This structure is therefore not merely descriptive; it provides information that is useful for anticipating agreement under changes in the evaluation set.
Fidelity is better understood as conditional on regions of the evaluation space, not only as a scalar average.

However, localizing fidelity improves characterization without changing what fidelity can establish---it remains a measure of observational agreement.
The matched Breast Cancer surrogates make this limitation concrete: identical global and boundary-conditioned fidelity can coexist with sharply different counterfactual transfer behavior.
This distinction is important because a surrogate may reproduce the teacher's observed predictions while responding differently to changes selected using the surrogate itself.
Boundary-aware fidelity should therefore be interpreted as additional evidence about the teacher--surrogate relationship rather than as a guarantee that the surrogate will preserve every downstream behavior of interest.
Which forms of agreement matter will ultimately depend on the intended use of the surrogate.

A further complication is that the regions used for boundary-aware evaluation may themselves depend on the teacher being evaluated.
Our deep-network experiments show that independently trained teachers can exhibit very high global predictive agreement while identifying substantially different examples as belonging to their boundary regions.
Thus, ``near the decision boundary'' should not necessarily be treated as a model-independent property of the evaluation space.
Boundary-aware fidelity in deep networks is therefore best interpreted relative to a particular teacher unless boundary stability across training runs has been established.

These conclusions should be interpreted within the scope of our experiments. In particular:

\begin{itemize}
    \item \textbf{Empirical scope.} Our experiments cover a limited set of datasets and model families. The tabular analyses use several benchmark datasets, while deep boundary stability is examined only on MNIST and CIFAR-10 with a finite set of independently trained teachers.
    
    \item \textbf{Operational scope.} Our segment-crossing measure is an approximate geometric proxy rather than an exact distance to the decision boundary, and our evaluation-composition experiments manipulate the composition of the evaluation set rather than claiming to capture arbitrary real-world distribution shifts.
    
    \item \textbf{Downstream scope.} Our counterfactual-transfer experiment is example-based and measures whether surrogate-selected changes transfer to the teacher; it should not be interpreted as establishing causal or actionable recourse.
\end{itemize}

\section{Conclusion}
\label{sec:conc}

Across several tabular datasets, surrogate families, and capacities, we find that teacher--surrogate disagreement is often concentrated near the teacher's decision boundary under both confidence-based and approximate segment-crossing definitions of boundary proximity.
This structure is not merely descriptive: confidence-stratified fidelity profiles better predict fidelity when evaluation composition changes than the corresponding global score.

At the same time, boundary-aware fidelity is not a complete characterization of the teacher--surrogate relationship.
Surrogates with similar global and boundary-local fidelity can exhibit sharply different example-based counterfactual transfer behavior, and independently trained deep teachers with high predictive agreement need not identify the same evaluation examples as belonging to their boundary regions.
Therefore, fidelity should be interpreted as conditional on a teacher--surrogate pair and its evaluation space, rather than as a single definitive measure of behavioral equivalence.

Ultimately, for surrogate fidelity, we argue that alongside the question posed by \citet{ribeiroWhyShouldTrust2016}, ``\textit{Why should I trust you?}'', we must also ask: ``\textit{Where can I trust you?}''

\subsection*{Reproducibility Statement}
Experimental code is available at \url{https://github.com/jacksoneshbaugh/boundary-aware-surrogate-fidelity/tree/master}.

\subsection*{Acknowledgments}
I thank Professors Jorge Silveyra and Sofia Serrano for their guidance and feedback on this work. I am also deeply grateful to Professors Silveyra, Serrano, and Jeffrey Pfaffmann for the mentorship that has shaped my undergraduate education. I also thank the Lafayette College Department of Computer Science for its continued support of my work.

\printbibliography

\appendix

\section{Experimental Details}
\label{app:experimental_details}

This appendix provides implementation details for the experiments reported in Sections~\ref{sec:boundary_results}--\ref{sec:deep_results}.
Unless otherwise stated, tabular experiments are repeated over ten random seeds.
All feature-standardization statistics are estimated using the corresponding training partition only.

\subsection{Datasets and Preprocessing}
\label{app:datasets}

Our tabular experiments use four datasets distributed with \texttt{scikit-learn}, each interpreted as a binary classification task for the purposes of this study.
Table~\ref{tab:datasets} summarizes these tasks.

\begin{table}[H]
    \centering
    \begin{tabular}{lrrl}
        \toprule
        Dataset & Examples & Features & Binary task \\
        \midrule
        Breast Cancer Wisconsin~\citep{wolberg1993wdbc} & 569 & 30 & Original binary labels \\
        Wine~\citep{aeberhard1992wine} & 178 & 13 & Class 0 vs.\ remaining classes \\
        Iris~\citep{fisher1936iris} & 150 & 4 & Class 0 vs.\ remaining classes \\
        Diabetes~\citep{efron2004least} & 442 & 10 & Target $\geqslant$ median vs.\ target $<$ median \\
        \bottomrule
    \end{tabular}
    \caption{Tabular datasets used in the experiments and the binary classification tasks constructed from them.
    Wine and Iris are converted to one-vs-rest tasks using class 0 as the positive class.
    The Diabetes target is binarized at its median.}
    \label{tab:datasets}
\end{table}

For each random seed, data partitions are generated using stratified sampling with respect to the binary ground-truth label.
Features are standardized using a \texttt{StandardScaler} fit only on the training partition, and the resulting transformation is then applied unchanged to all held-out partitions.
Experiment-specific partition sizes are described below.

\subsection{Tabular Teacher Models}

For all tabular experiments, the teacher is a binary multilayer perceptron with two hidden layers of widths 64 and 32, ReLU activations, and a single scalar output logit.
Teachers are trained on the ground-truth binary labels using binary cross-entropy with logits and Adam with a learning rate \( 10^{-2} \).
Training is full-batch for 500 epochs.
Predictions are obtained by thresholding the output logit at zero.

A fresh teacher is trained for each random seed.
Before model construction and training, we seed Python, NumPy, and PyTorch random number generators, including CUDA when available.
Teacher confidence in the binary experiments is measured using absolute logit magnitude, \( c_T(x)=|f_T(x)| \).

\subsection{Surrogate Models}

All surrogates are trained on the training inputs using the corresponding teacher hard predictions as targets, rather than the ground-truth labels.
Table~\ref{tab:surrogates} summarizes the representative surrogate configurations used in the cross-family analyses.

\begin{table}[H]
    \centering
    \begin{tabular}{ll}
        \toprule
        Surrogate family & Configuration \\
        \midrule
        Decision tree & maximum depth \( 5 \) \\
        Logistic regression & \( C=1 \), \texttt{liblinear} solver \\
        \( k \)-NN & \( k=5 \) \\
        MLP & one hidden layer, width \( 32 \), ReLU, \texttt{lbfgs} \\
        \bottomrule
    \end{tabular}
    \caption{Representative surrogate configurations used in the cross-family robustness analyses.}
    \label{tab:surrogates}
\end{table}

Our primary boundary-fidelity capacity analysis varies decision-tree maximum depth over \( \{1,2,3,5,7,10\} \).
The fidelity-prediction analysis uses depths \( \{3,5,7\} \), while the counterfactual-transfer analysis uses \( \{2,3,5,7,10\} \).
Cross-family robustness analyses use the representative configurations in Table~\ref{tab:surrogates}.
Logistic regression and MLP surrogates are fit with a maximum of 5000 iterations; all other parameters use the corresponding \texttt{scikit-learn} defaults unless otherwise stated.
Surrogate models with stochastic components use the experimental seed.

\subsection{Boundary-Proximity Computation}

For confidence-based proximity, we compute the absolute teacher logit \( c_T(x)=|f_T(x)| \) for each evaluation example.
For a requested boundary fraction \( \alpha \), the boundary region contains the \( \lceil \alpha |X| \rceil \) examples with the smallest confidence values.

For the segment-crossing proxy, each evaluation example \( x \) is paired with the nearest observed evaluation example \( x' \) receiving the opposite teacher prediction, where distance is Euclidean in the standardized feature space.
We then perform 30 iterations of binary search along the line segment joining \( x \) and \( x' \) to approximate a teacher decision crossing.
Segment-crossing proximity is the Euclidean distance from \( x \) to this estimated crossing.
This quantity is an operational geometric proxy and is not the exact minimum Euclidean distance from \( x \) to the teacher's full decision boundary.

If no oppositely predicted evaluation example exists, the segment-crossing distance is treated as undefined and the example is excluded from the valid geometric ranking.
For segment-crossing proximity, the requested boundary fraction is applied to this valid ranking.
Boundary regions are evaluated at \( \alpha\in\{0.10,0.20,0.25,0.40\} \).
For a fixed teacher, random seed, and evaluation set, confidence-based and segment-crossing boundary regions are computed independently of the surrogate and reused across all surrogate configurations.

\subsection{Deep Teacher Models}

The deep-teacher stability analysis uses five independently trained teachers for each of MNIST~\citep{lecun1998gradient} and CIFAR-10~\citep{krizhevsky2009learning}, corresponding to random seeds \( \{0,1,2,3,4\} \).

For MNIST, we use a small convolutional network with three \( 3\times3 \) convolutional layers with stride 1 and padding 1, containing 32, 64, and 128 channels.
The first two convolutional layers are followed by \( 2\times2 \) max-pooling.
The convolutional representation is then mapped through a fully connected layer of width 256 with ReLU activation and dropout \( 0.5 \), followed by a 10-class output layer.
Images are normalized using mean \( 0.1307 \) and standard deviation \( 0.3081 \).
Models are trained for 30 epochs using cross-entropy loss and Adam with learning rate \( 10^{-3} \).

For CIFAR-10, we use ResNet-18 models trained from scratch.
To accommodate \( 32\times32 \) inputs, the initial convolution is replaced with a \( 3\times3 \) convolution with stride 1 and padding 1, and the initial max-pooling layer is removed.
Training data are augmented using random \( 32\times32 \) crops with four pixels of padding and random horizontal flips.
Images are normalized using channel means \( (0.4914, 0.4822, 0.4465) \) and standard deviations \( (0.2023, 0.1994, 0.2010) \).
Models are trained for 200 epochs using cross-entropy loss and SGD with learning rate \( 0.1 \), momentum \( 0.9 \), weight decay \( 5\times10^{-4} \), and cosine learning-rate annealing.

Both datasets use a batch size of 128.
Python, NumPy, and PyTorch random number generators are seeded independently for each run; CUDA seeds are also set when available, and cuDNN deterministic execution is enabled.
For the deep-teacher experiments, boundary proximity is operationalized using the margin between the largest and second-largest output logits, with smaller margins indicating greater proximity to the decision boundary.
\section{Additional Boundary-Fidelity Results}
\label{app:boundary_results}

This section provides additional numerical detail for the boundary-local fidelity analysis in Section~\ref{sec:boundary_results}.
We first report the decision-tree capacity sweep underlying the main boundary-local results, then extend the surrogate-family comparison across all evaluated boundary fractions.

\subsection{Decision-Tree Capacity Sweep}
\label{app:boundary_capacity}

Table~\ref{tab:capacity_sweep} reports the numerical results underlying the decision-tree capacity analysis in Section~\ref{sec:boundary_results}.
Values are averaged over ten random seeds.
We report the \( 25\% \) boundary region as a representative operating point; Figure~\ref{fig:boundary_fidelity} in the main text shows how the global-to-boundary gap varies across all evaluated boundary fractions.

\begin{table}[H]
    \centering
    \begin{tabular}{lrrrr}
        \toprule
        Dataset & Depth & Global fidelity (\%) & Confidence gap (pp) & Segment-crossing gap (pp) \\
        \midrule
        Breast Cancer & 1  & 91.6 & 12.3 $\pm$ 3.2 & 13.7 $\pm$ 2.5 \\
                      & 2  & 93.3 & 11.2 $\pm$ 3.0 & 10.8 $\pm$ 3.9 \\
                      & 3  & 93.4 & 11.8 $\pm$ 2.4 & 11.8 $\pm$ 3.0 \\
                      & 5  & 94.0 & 11.9 $\pm$ 2.2 & 10.1 $\pm$ 2.8 \\
                      & 7  & 94.1 & 12.9 $\pm$ 2.2 & 10.6 $\pm$ 2.4 \\
                      & 10 & 94.2 & 12.5 $\pm$ 2.0 & 10.9 $\pm$ 2.6 \\
        \addlinespace
        Diabetes      & 1  & 72.5 & 17.5 $\pm$ 5.8 & 16.0 $\pm$ 6.3 \\
                      & 2  & 75.0 & 18.6 $\pm$ 4.1 & 15.9 $\pm$ 4.4 \\
                      & 3  & 74.1 & 18.8 $\pm$ 4.7 & 13.8 $\pm$ 4.0 \\
                      & 5  & 73.6 & 18.6 $\pm$ 4.2 & 13.9 $\pm$ 4.6 \\
                      & 7  & 74.3 & 19.9 $\pm$ 3.4 & 15.2 $\pm$ 3.6 \\
                      & 10 & 72.4 & 19.8 $\pm$ 2.9 & 13.6 $\pm$ 3.9 \\
        \addlinespace
        Wine          & 1  & 90.6 & 7.0 $\pm$ 4.7  & 4.8 $\pm$ 5.7 \\
                      & 2  & 95.4 & 10.4 $\pm$ 5.5 & 8.9 $\pm$ 5.2 \\
                      & 3  & 95.0 & 10.0 $\pm$ 5.8 & 10.7 $\pm$ 4.9 \\
                      & 5  & 95.0 & 11.4 $\pm$ 5.2 & 11.4 $\pm$ 4.4 \\
                      & 7  & 95.0 & 11.4 $\pm$ 5.2 & 11.4 $\pm$ 4.4 \\
                      & 10 & 95.0 & 11.4 $\pm$ 5.2 & 11.4 $\pm$ 4.4 \\
        \addlinespace
        Iris          & 1  & 99.6 & 1.2 $\pm$ 1.6 & 1.2 $\pm$ 1.6 \\
                      & 2  & 99.6 & 1.2 $\pm$ 1.6 & 1.2 $\pm$ 1.6 \\
                      & 3  & 99.6 & 1.2 $\pm$ 1.6 & 1.2 $\pm$ 1.6 \\
                      & 5  & 99.6 & 1.2 $\pm$ 1.6 & 1.2 $\pm$ 1.6 \\
                      & 7  & 99.6 & 1.2 $\pm$ 1.6 & 1.2 $\pm$ 1.6 \\
                      & 10 & 99.6 & 1.2 $\pm$ 1.6 & 1.2 $\pm$ 1.6 \\
        \bottomrule
    \end{tabular}
    \caption{Decision-tree capacity sweep at boundary fraction \( \alpha=0.25 \).
    Global fidelity and boundary gaps are averaged over ten random seeds.
    Boundary gaps are defined as \( F_{\mathrm{global}}-F_{\mathrm{boundary}} \); gap values are shown as means with approximate 95\% confidence intervals.}
    \label{tab:capacity_sweep}
\end{table}

\subsection{Surrogate-Family Robustness}
\label{app:boundary_families}

\begin{figure}[H]
    \centering
    \includegraphics[width=0.8\linewidth]{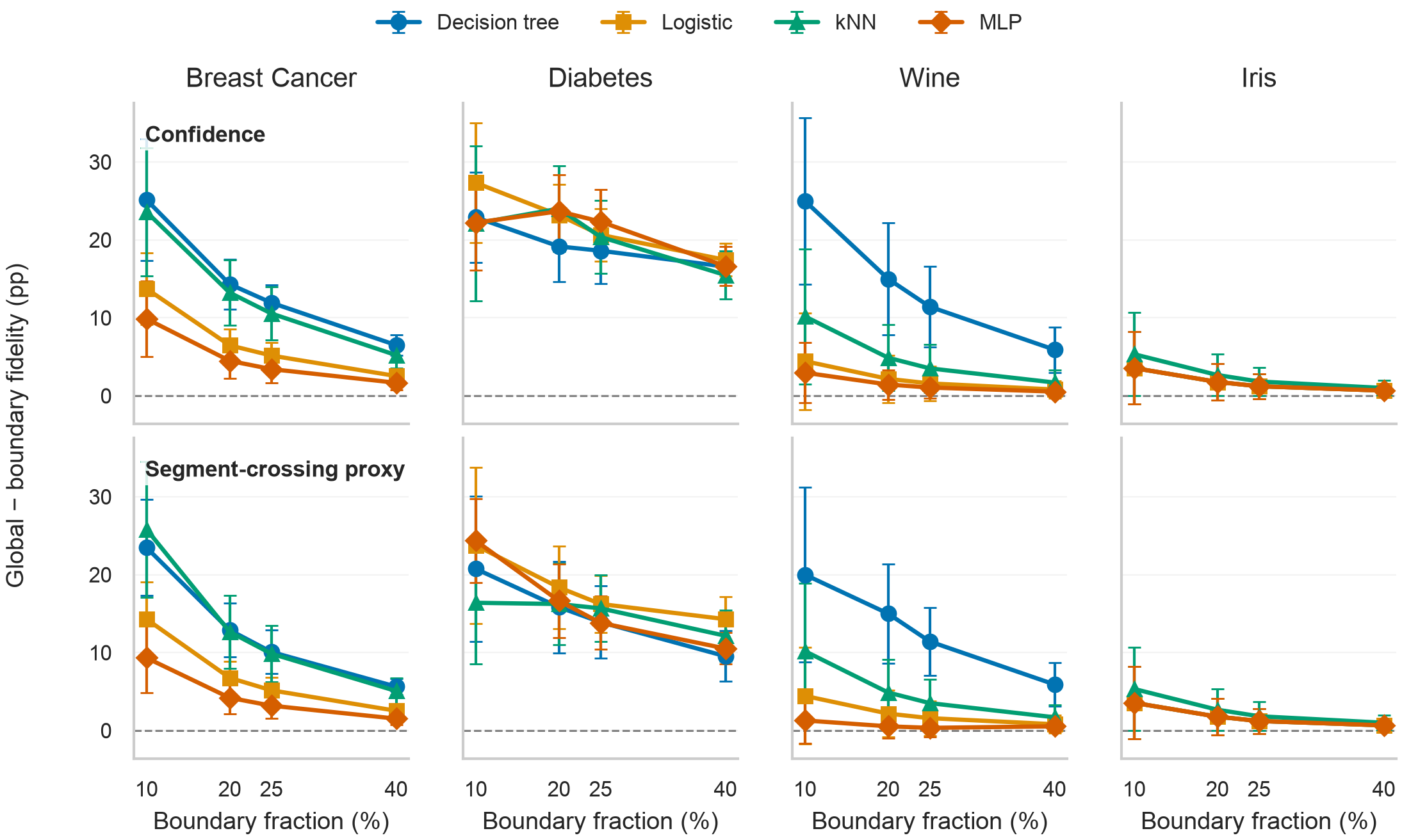}
    \caption{Surrogate-family robustness across boundary fractions.
    Global-to-boundary fidelity gaps are shown for representative decision-tree, logistic-regression, \( k \)-NN, and MLP surrogates under confidence-based (top) and approximate segment-crossing (bottom) measures of boundary proximity.
    Points show means over ten random seeds; error bars indicate approximate 95\% confidence intervals.}
    \label{fig:family_boundary_fractions}
\end{figure}

Table~\ref{tab:surrogate_family_robustness} in the main text compares representative surrogate families at a boundary fraction of \( 25\% \).
Figure~\ref{fig:family_boundary_fractions} extends this analysis across all evaluated boundary fractions.
The qualitative pattern is stable as the boundary fraction varies: Breast Cancer and Diabetes exhibit positive boundary-local gaps across all four surrogate families under both boundary definitions.
On Wine, the effect is smaller for the non-tree surrogates, whose global fidelity is already near saturation, while Iris remains a near-null case throughout.

\section{Additional Fidelity-Prediction Results}
\label{app:shift_results}

This section provides robustness analyses for the fidelity-prediction experiment in Section~\ref{sec:shift_results}.
We first examine sensitivity to the number of confidence bins used to construct the fidelity profile, then test whether the improvement over global fidelity persists across representative surrogate families.

\subsection{Five-Bin Sensitivity}
\label{app:shift_five_bins}

Our primary fidelity-prediction analysis constructs the conditional fidelity profile using three confidence bins.
Table~\ref{tab:shift_bins} compares this choice with a five-bin profile using the same source and target partitions.
If a source confidence bin contains no examples, its conditional fidelity is set to the source-set global fidelity for purposes of constructing the profile prediction.
Overall, the five-bin profile retains an advantage over the global-fidelity baseline, although its mean absolute error is slightly higher than that of the three-bin profile.
The difference is concentrated on Wine, the smallest dataset; Breast Cancer is essentially unchanged, while Diabetes shows only a small increase in error.
The corresponding overall reduction in mean absolute error relative to the global baseline is \( 39.7\% \) with three bins and \( 34.6\% \) with five bins.

\begin{table}[H]
    \centering
    \begin{tabular}{lrrr}
        \toprule
        Dataset & Global baseline (pp) & 3-bin profile (pp) & 5-bin profile (pp) \\
        \midrule
        Breast Cancer & 4.30 & 2.93 & 2.92 \\
        Diabetes      & 13.62 & 5.92 & 6.04 \\
        Wine          & 5.09 & 5.02 & 6.10 \\
        \midrule
        Overall       & 7.67 & 4.62 & 5.02 \\
        \bottomrule
    \end{tabular}
    \caption{Sensitivity of target-fidelity prediction to the number of confidence bins.
    Values are mean absolute prediction errors across surrogate depths, random seeds, and target compositions.
    The global-fidelity baseline is unchanged by the number of profile bins.}
    \label{tab:shift_bins}
\end{table}

\subsection{Surrogate-Family Robustness}
\label{app:shift_families}

We additionally repeat the three-bin fidelity-prediction analysis using representative decision-tree, logistic-regression, \( k \)-NN, and MLP surrogates.
Table~\ref{tab:shift_families} reports results overall and separately by dataset.
Averaged across datasets and target compositions, the confidence-stratified profile has lower mean absolute prediction error for every surrogate family.
The improvement is especially pronounced on Diabetes.
Breast Cancer also improves for three of the four families, while the MLP configuration is effectively unchanged; on Wine, several high-fidelity surrogates produce little or no difference between the two predictors.

\begin{table}[H]
    \centering
    \begin{tabular}{llrrr}
        \toprule
        Scope & Surrogate & Global baseline (pp) & 3-bin profile (pp) & Improvement (pp) \\
        \midrule
        Overall
            & Decision tree & 8.00 & 5.06 & 2.95 \\
            & Logistic      & 6.78 & 2.96 & 3.82 \\
            & \( k \)-NN    & 7.26 & 4.07 & 3.19 \\
            & MLP           & 5.13 & 2.81 & 2.32 \\
        \addlinespace
        Breast Cancer
            & Decision tree & 4.56 & 3.63 & 0.93 \\
            & Logistic      & 1.93 & 1.44 & 0.49 \\
            & \( k \)-NN    & 4.39 & 3.14 & 1.25 \\
            & MLP           & 1.23 & 1.29 & -0.06 \\
        \addlinespace
        Diabetes
            & Decision tree & 14.17 & 6.33 & 7.84 \\
            & Logistic      & 17.59 & 6.62 & 10.97 \\
            & \( k \)-NN    & 14.06 & 6.07 & 7.99 \\
            & MLP           & 13.33 & 6.30 & 7.03 \\
        \addlinespace
        Wine
            & Decision tree & 5.28 & 5.21 & 0.07 \\
            & Logistic      & 0.83 & 0.83 & 0.00 \\
            & \( k \)-NN    & 3.33 & 2.99 & 0.35 \\
            & MLP           & 0.83 & 0.83 & 0.00 \\
        \bottomrule
    \end{tabular}
    \caption{Fidelity-prediction robustness across surrogate families.
    Values are mean absolute target-fidelity prediction errors.
    ``Improvement'' is computed before rounding as global baseline error minus profile error, so positive values favor the confidence-stratified profile.}
    \label{tab:shift_families}
\end{table}
\section{Additional Counterfactual-Transfer Results}
\label{app:counterfactual_results}

This section provides additional detail for the example-based counterfactual-transfer analysis in Section~\ref{sec:counterfactual_results}.
We first report results across all evaluated surrogate capacities, then examine specific surrogate pairs with closely matched observational fidelity but substantially different transfer behavior.

\subsection{Full Configuration Results}
\label{app:counterfactual_full}

Table~\ref{tab:counterfactual_full} reports results for every evaluated decision-tree depth, averaged over ten random seeds.
Global and boundary-conditioned fidelity can remain relatively stable across surrogate capacities even when counterfactual transfer rates differ.
Across all configurations, the experiment contains 10,250 eligible transfer attempts; every eligible example has an available candidate with the opposite surrogate prediction.

\begin{table}[H]
    \centering
    \begin{tabular}{lrrrr}
        \toprule
        Dataset & Depth & Global fidelity (\%) & Boundary fidelity (\%) & Transfer rate (\%) \\
        \midrule
        Breast Cancer
            & 2  & 92.4 $\pm$ 2.1 & 79.7 $\pm$ 4.1 & 50.2 $\pm$ 12.5 \\
            & 3  & 93.8 $\pm$ 1.7 & 83.1 $\pm$ 3.8 & 51.9 $\pm$ 11.1 \\
            & 5  & 94.6 $\pm$ 1.5 & 84.8 $\pm$ 4.6 & 49.0 $\pm$ 14.2 \\
            & 7  & 94.3 $\pm$ 1.7 & 84.5 $\pm$ 3.9 & 45.3 $\pm$ 13.5 \\
            & 10 & 94.5 $\pm$ 1.7 & 84.5 $\pm$ 4.4 & 44.4 $\pm$ 13.9 \\
        \addlinespace
        Diabetes
            & 2  & 73.9 $\pm$ 2.5 & 48.2 $\pm$ 5.5 & 50.2 $\pm$ 11.1 \\
            & 3  & 73.4 $\pm$ 4.0 & 45.9 $\pm$ 4.9 & 47.4 $\pm$ 8.9 \\
            & 5  & 71.4 $\pm$ 2.5 & 49.5 $\pm$ 5.9 & 46.1 $\pm$ 8.1 \\
            & 7  & 71.2 $\pm$ 2.1 & 52.3 $\pm$ 8.3 & 46.4 $\pm$ 8.3 \\
            & 10 & 71.7 $\pm$ 1.2 & 54.1 $\pm$ 7.4 & 48.2 $\pm$ 8.5 \\
        \addlinespace
        Wine
            & 2  & 96.4 $\pm$ 1.8 & 88.9 $\pm$ 4.6 & 69.2 $\pm$ 15.3 \\
            & 3  & 95.0 $\pm$ 1.6 & 82.2 $\pm$ 5.8 & 72.2 $\pm$ 14.3 \\
            & 5  & 95.0 $\pm$ 1.6 & 82.2 $\pm$ 5.8 & 68.3 $\pm$ 13.0 \\
            & 7  & 95.0 $\pm$ 1.6 & 82.2 $\pm$ 5.8 & 68.3 $\pm$ 13.0 \\
            & 10 & 95.0 $\pm$ 1.6 & 82.2 $\pm$ 5.8 & 68.3 $\pm$ 13.0 \\
        \bottomrule
    \end{tabular}
    \caption{Counterfactual-transfer results across decision-tree capacities.
    Fidelity and transfer rates are reported as means across ten random seeds, with \( \pm \) denoting approximate 95\% confidence-interval half-widths.
    Boundary-local fidelity uses \( \alpha=0.25 \).}
    \label{tab:counterfactual_full}
\end{table}

\subsection{Matched-Fidelity Examples}
\label{app:counterfactual_matched}

We define a matched pair as two surrogate configurations trained under the same dataset and random seed whose global fidelity and boundary-conditioned fidelity each differ by at most one percentage point.
Among qualifying pairs, Table~\ref{tab:counterfactual_matched} reports the pair with the largest counterfactual-transfer gap for each dataset.
For Breast Cancer, the depth-3/depth-7 pair is also the example highlighted in Figure~\ref{fig:counterfactual_transfer}.

\begin{table}[H]
    \centering
    \begin{tabular}{lrrrrrr}
        \toprule
        Dataset & Seed & Depths & Global fidelity (\%) & Boundary fidelity (\%)
                & Transfer rates (\%) & Gap (pp) \\
        \midrule
        Breast Cancer
            & 3 & 3 / 7
            & 94.7 / 94.7
            & 82.8 / 82.8
            & 73.1 / 24.1
            & 49.1 \\
        Diabetes
            & 7 & 5 / 10
            & 72.7 / 72.7
            & 54.5 / 54.5
            & 56.2 / 37.5
            & 18.8 \\
        Wine
            & 4 & 3 / 10
            & 100.0 / 100.0
            & 100.0 / 100.0
            & 100.0 / 61.1
            & 38.9 \\
        \bottomrule
    \end{tabular}
    \caption{Matched-fidelity surrogate pairs with differing counterfactual-transfer behavior.
    Each row compares two decision-tree surrogates trained under the same dataset and random seed.
    The transfer gap is the difference between the two transfer rates in percentage points.}
    \label{tab:counterfactual_matched}
\end{table}

These examples make the limitation of observational fidelity particularly clear.
In each case, the matched surrogates have the same global and boundary-conditioned fidelity at the reported precision, yet changes selected using the surrogate transfer to the teacher at substantially different rates.
The Wine example is especially stark: both surrogates achieve perfect global and boundary-conditioned fidelity, while their transfer rates differ by nearly 39 percentage points.
Thus, even highly localized observational agreement does not uniquely determine behavior under surrogate-selected changes.
\section{Additional Deep-Teacher Stability Results}
\label{app:deep_results}

This section provides additional numerical detail for the deep-teacher stability analysis in Section~\ref{sec:deep_results}.
We first report the predictive accuracies of the independently trained teachers, then provide the pairwise stability statistics underlying Figure~\ref{fig:deep_boundary_stability}.

\subsection{Teacher Accuracies}
\label{app:deep_accuracies}

Table~\ref{tab:deep_teacher_accuracies} reports test accuracy for each independently trained teacher.
The teachers within each dataset achieve closely matched overall predictive performance, providing a comparable basis for the pairwise boundary-stability analysis.

\begin{table}[H]
    \centering
    \begin{tabular}{lrrrrr}
        \toprule
        Dataset & Seed 0 & Seed 1 & Seed 2 & Seed 3 & Seed 4 \\
        \midrule
        MNIST
            & 99.09 & 99.27 & 99.33 & 99.18 & 99.39 \\
        CIFAR-10
            & 95.30 & 95.27 & 95.35 & 95.42 & 95.63 \\
        \bottomrule
    \end{tabular}
    \caption{Test accuracy (\%) of independently trained deep teachers.
    Each teacher is evaluated on the common 10,000-example test set for its dataset.}
    \label{tab:deep_teacher_accuracies}
\end{table}

\subsection{Pairwise Boundary-Set Stability}
\label{app:deep_pairwise}

For each dataset, we compare all ten unique pairs among the five independently trained teachers.
Table~\ref{tab:deep_pairwise} reports the mean pairwise statistics at each evaluated boundary fraction \( \alpha \).
Global agreement is computed over the full test set.
Boundary-set overlap is measured by Jaccard similarity, while union and intersection agreement measure predictive agreement on the union and intersection, respectively, of the two teachers' boundary sets.

\begin{table}[H]
    \centering
    \begin{tabular}{lrrrrrr}
        \toprule
        Dataset & $\alpha$
        & Global agr. (\%)
        & Jaccard (\%)
        & Union agr. (\%)
        & Intersection agr. (\%)
        & Global $-$ union (pp) \\
        \midrule
        MNIST
            & 0.10 & 99.18 & 39.39 & 94.31 & 86.63 & 4.87 \\
            & 0.20 & 99.18 & 43.13 & 97.08 & 93.32 & 2.10 \\
            & 0.25 & 99.18 & 45.01 & 97.63 & 94.77 & 1.55 \\
        \addlinespace
        CIFAR-10
            & 0.10 & 96.05 & 44.32 & 72.09 & 58.39 & 23.96 \\
            & 0.20 & 96.05 & 56.81 & 84.50 & 74.63 & 11.54 \\
            & 0.25 & 96.05 & 58.53 & 87.47 & 79.50 & 8.58 \\
        \bottomrule
    \end{tabular}
    \caption{Pairwise stability of independently trained deep teachers.
    Values are means across the ten unique teacher pairs for each dataset.}
    \label{tab:deep_pairwise}
\end{table}

The numerical results reinforce the distinction between global predictive agreement and stability of the examples identified as lying near the decision boundary.
At the \( 10\% \) boundary fraction, pairwise global agreement is \( 99.18\% \) on MNIST and \( 96.05\% \) on CIFAR-10, while mean boundary-set Jaccard overlap is only \( 39.39\% \) and \( 44.32\% \), respectively.
Predictive agreement also falls within these regions, particularly on CIFAR-10, where agreement on the union of the two teachers' boundary sets is \( 72.09\% \).
As the boundary fraction increases, overlap and boundary-region agreement increase, but independently trained teachers continue to identify substantially different sets of examples as lying near the decision boundary despite their high global predictive agreement.

\end{document}